\documentclass[sigconf]{acmart}

\AtBeginDocument{%
  \providecommand\BibTeX{{%
    \normalfont B\kern-0.5em{\scshape i\kern-0.25em b}\kern-0.8em\TeX}}}

\copyrightyear{2023} 
\acmYear{2023} 
\setcopyright{acmlicensed}\acmConference[WSDM '23]{Proceedings of the Sixteenth ACM International Conference on Web Search and Data Mining}{February 27-March 3, 2023}{Singapore, Singapore}
\acmBooktitle{Proceedings of the Sixteenth ACM International Conference on Web Search and Data Mining (WSDM '23), February 27-March 3, 2023, Singapore, Singapore}
\acmPrice{15.00}
\acmDOI{10.1145/3539597.3570486}
\acmISBN{978-1-4503-9407-9/23/02}

\usepackage{amsmath,amsfonts,bm}

\def\eqref#1{equation~\ref{#1}}
\def\1{\bm{1}}

\def\vzero{{\bm{0}}}
\def\vone{{\bm{1}}}

\def\vc{{\bm{c}}}

\def\ve{{\bm{e}}}

\def\vx{{\bm{x}}}

\DeclareMathAlphabet{\mathsfit}{\encodingdefault}{\sfdefault}{m}{sl}
\SetMathAlphabet{\mathsfit}{bold}{\encodingdefault}{\sfdefault}{bx}{n}

\def\sA{{\mathbb{A}}}

\def\sS{{\mathbb{S}}}

\newcommand{\E}{\mathbb{E}}

\newcommand{\R}{\mathbb{R}}

\DeclareMathOperator*{\argmax}{arg\,max}

\usepackage{amsmath}  % for "align" environment
\usepackage{amsthm}   % for "proof" environment
\usepackage{amsfonts} % for \mathbb{}
\usepackage{dsfont}   % for indicator function font
\newtheorem{example}{Example}
\newtheorem{theorem}{Theorem}[section]

\newtheorem{corollary}[theorem]{Corollary} % Counter reset every time a new theorem environment is used.
\newtheorem{proposition}[theorem]{Proposition} % 
\usepackage{algorithm} % for algorithm env
\usepackage{algorithmic} % for algorithm env
\renewcommand{\algorithmiccomment}[1]{\bgroup\hfill//~#1\egroup} % for comment in algorithm env

\usepackage{pifont} % cross and tick
\usepackage{array} % using m{5em} in table
\newcolumntype{L}[1]{>{\raggedright\arraybackslash}m{#1}}  %% left aligned
\newcolumntype{C}[1]{>{\centering\arraybackslash}m{#1}} % for center and middle align in tabular
\newcolumntype{R}[1]{>{\raggedleft\arraybackslash}m{#1}}  %% right aligned
\usepackage{tablefootnote} % tablefootnote

\usepackage{graphicx}

\usepackage{subcaption} % subfigure environment

\usepackage{printlen} % to print line width: \printlength\textwidth

\def\sA{{\mathcal{A}}}

\def\sS{{\mathcal{S}}}

\def\conv{{\mathtt{conv}}} % convex hull
\def\aff{{\mathtt{aff}}} % affine hull
\def\valpha{{\boldsymbol{\alpha}}}

\def\vlambda{{\boldsymbol{\lambda}}}

\def\vtau{{\boldsymbol{\tau}}}

\usepackage{systeme,mathtools}
\usepackage{multirow} % use multirow in table
\usepackage{makecell} % use makecell in table
\usepackage{footnote}

\usepackage{balance}

\begin{document}

%%
%% The "title" command has an optional parameter,
%% allowing the author to define a "short title" to be used in page headers.

% Sequential incentive allocations.

\title[Marketing Budget Allocation with Offline Constrained Deep Reinforcement Learning]{Marketing Budget Allocation with Offline Constrained \\Deep Reinforcement Learning}
% Offline Constrained Reinforcement Learning \\ for Marketing Budget Allocation 

%%
%% The "author" command and its associated commands are used to define
%% the authors and their affiliations.
%% Of note is the shared affiliation of the first two authors, and the
%% "authornote" and "authornotemark" commands
%% used to denote shared contribution to the research.

\author{Tianchi Cai}
\affiliation{
    \institution{Ant Group}
    \country{Hangzhou, China}
}
\authornote{Equal contribution.}
\authornote{Corresponding author. <tianchi.ctc@antgroup.com>}

\author{Jiyan Jiang}
\affiliation{
    \institution{Tsinghua University}
    \country{Beijing, China}
}
\authornotemark[1]

\author{Wenpeng Zhang}
\affiliation{
    \institution{Ant Group}
    \country{Beijing, China}
}

\author{Shiji Zhou}
\affiliation{
    \institution{Tsinghua University}
    \country{Beijing, China}
}

\author{Xierui Song}
\affiliation{
    \institution{Ant Group}
    \country{Hangzhou, China}
}

\author{Li Yu}
\affiliation{
    \institution{Ant Group}
    \country{Hangzhou, China}
}

\author{Lihong Gu}
\affiliation{
    \institution{Ant Group}
    \country{Hangzhou, China}
}

\author{Xiaodong Zeng}
\affiliation{
    \institution{Ant Group}
    \country{Hangzhou, China}
}

\author{Jinjie Gu}
\affiliation{
    \institution{Ant Group}
    \country{Hangzhou, China}
}

\author{Guannan Zhang}
\affiliation{
    \institution{Ant Group}
    \country{Hangzhou, China}
}

%\author{Wenpeng Zhang, Shiji Zhou, Xierui Song, Li Yu, Lihong Gu, Xiaodong Zeng, Jinjie Gu, Guannan Zhang}

% \email{tianchi.ctc@antgroup.com}
% \affiliation{%
%   \institution{Ant Group, China}
% }
% 

%%
%% By default, the full list of authors will be used in the page
%% headers. Often, this list is too long, and will overlap
%% other information printed in the page headers. This command allows
%% the author to define a more concise list
%% of authors' names for this purpose.
\renewcommand{\shortauthors}{Tianchi Cai et al.}

%%
%% The abstract is a short summary of the work to be presented in the
%% article.
\begin{abstract}
We study the budget allocation problem in online marketing campaigns that utilize previously collected offline data. We first discuss the long-term effect of optimizing marketing budget allocation decisions in the offline setting. To overcome the challenge, we propose a novel game-theoretic offline value-based reinforcement learning method using mixed policies. The proposed method reduces the need to store infinitely many policies in previous methods to only constantly many policies, which achieves nearly optimal policy efficiency, making it practical and favorable for industrial usage. We further show that this method is guaranteed to converge to the optimal policy, which cannot be achieved by previous value-based reinforcement learning methods for marketing budget allocation. Our experiments on a large-scale marketing campaign with tens-of-millions users and more than one billion budget verify the theoretical results and show that the proposed method outperforms various baseline methods. The proposed method has been successfully deployed to serve all the traffic of this marketing campaign.
\end{abstract}

%%
%% The code below is generated by the tool at http://dl.acm.org/ccs.cfm.
%% Please copy and paste the code instead of the example below.
%%

% \ccsdesc[500]{Computing methodologies~Batch learning}

%%
%% Keywords. The author(s) should pick words that accurately describe
%% the work being presented. Separate the keywords with commas.

%%
%% This command processes the author and affiliation and title
%% information and builds there first part of the formatted document.

\maketitle

\section{Introduction}

\begin{figure*}[t] 
\begin{center}
\includegraphics[width=0.95\textwidth]{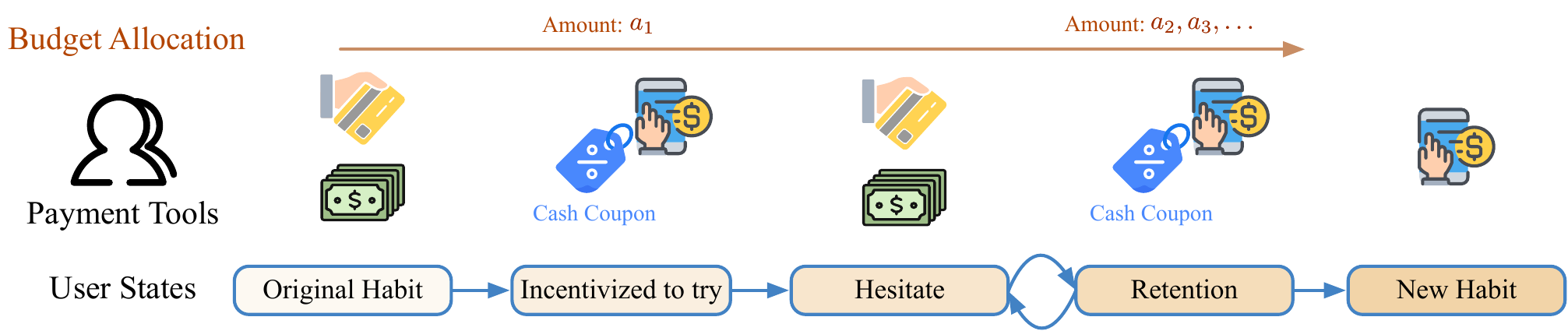}
\end{center}
\caption{In an online marketing campaign, cash coupons are offered to promote a digital payment app. Users can use one cash coupon every day, and the performance of the marketing campaign is measured by the average DAU over this period.
} \label{fig:caption}
\end{figure*} 

% Replace the ADAU by usually there is a budget constraint
Online marketing campaigns play a prominent role in user acquisition and retention. For example, as illustrated in Figure \ref{fig:caption}, E-commerce and digital payment companies often offer cash coupons to encourage purchases or promote new payment tools
\cite{zhao2019unified,yu2021joint,liu2019graph}. The core problem under these campaigns is appropriately allocating budgets to incentivize desired user behaviors while satisfying certain budget constraints. With the increasing scale of online marketing campaigns, the budget allocation problem has stimulated a great bulk of research attention.
  
The effectiveness of a marketing campaign is commonly measured by metrics regarding users' daily activeness, for example, the average Daily Active Users (DAU), which is the average of the DAU in a given time window (e.g. one week). As these metrics are hard to directly optimize, conventional methods use immediate user responses, like the coupon redemption rate \cite{yu2021joint}, as surrogates. Typically these methods take a two-step framework \cite{zhao2019unified,agarwal2014budget,xu2015smart,chen2021adversarial,li2021large}. They first build a response model, which estimates users' immediate responses to different incentives \cite{chen2021adversarial,liu2019graph}, then solve a constrained optimization problem to make the budget allocation \cite{agarwal2014budget,zhong2015stock}. Although this framework scales well, it fails to take the long-lasting effects of a marketing campaign into consideration. For example, new users may establish desired payment habits after participating in the marketing campaign several times, which cannot be modeled by response models using immediate signals.

Due to the ability to characterize the long-term effect, Reinforcement Learning (RL) has been introduced to the marketing budget allocation scenarios. For example, airline companies have used small-scale RL to optimize their frequent flyer program for a long time \cite{labbi2007optimizing}. With the recent development of constrained RL \cite{liang2018accelerated,paternain2019constrained,stooke2020responsive,chen2021primal}, various constrained policy optimization methods are proposed to optimize the allocation policy by interacting with user response simulators of the marketing campaign while satisfying required budget constraints \cite{chow2017risk,wu2018budget}.

However, accurate simulators are usually very difficult to build in industrial scenarios \cite{shi2019virtual,chen2021survey}. To eschew the need for simulators, various offline value-based methods (Q-learning) are proposed, which can learn merely with offline datasets. \citet{xiao2019model} proposes one of the most representative methods in this line of research which penalizes the Q-value function with constraint dissatisfaction. Although practically effective, it lacks a theoretical guarantee and may not converge to the optimal policy. To ensure the convergence of value-based methods, it is tempting to consider the recently proposed game-theoretic approaches that take advantage of mixed policies \cite{le2019batch,bohez2019value,miryoosefi2019reinforcement,zahavy2021reward}. A mixed policy can be viewed as a distribution over multiple individual policies. At the start of each episode, a single individual policy is randomly picked from these policies to generate the decision for the entire episode. Such methods are shown to converge as mixed policy contains (nearly) infinitely many individual policies as the number of training steps goes to infinity. However, in these mixed policy methods, the number of individual policies needed to be stored grows linearly, which makes them intractable to be implemented in industrial marketing scenarios where each individual policy is represented by a large-scale neural network as the function approximator.

In this paper, we formulate the marketing budget allocation problem as an offline constrained deep RL problem, in light of the failure of short-term decisions in capturing the long-term effects of marketing campaigns. Then we use a game-theoretic value-based method with mixed policies to solve this problem. To resolve the aforementioned issue of prohibitively high memory cost for storing all historical individual policies as in previous mixed policy methods \cite{le2019batch,miryoosefi2019reinforcement,zahavy2021reward}, we propose to express any mixed policy as a convex combination of a finite set of policies, called \textit{Affinely Independent Mixed}  (AIM) policies. These AIM policies form a basis of the affine space spanned by the policies' measurements (i.e., reward and budget constraints). Hence, this approach effectively reduces the number of stored individual policies to a constant (i.e., at most $m+2$ in the setting with $m$ constraints). We further devise two methods to determine the AIM policies and the weights during the learning process, i.e., AIM-mean and AIM-greedy. Specifically, AIM-mean recovers the original mixed policy methods \cite{le2019batch,miryoosefi2019reinforcement,zahavy2021reward} and converges to the optimal policy with constant memory cost. AIM-greedy is an analogy of the heuristic method with a single policy \cite{xiao2019model,le2019batch}, which outperforms the previous method empirically and is favorable for practical scenarios. Experiments in the real-world online marketing scenario verify our analysis and show superior empirical performance. 

The contributions of this paper are summarized as follows: \textbf{(i)} Using data collected from a real-world online marketing campaign, we reveal the long-term effect of the marketing campaign, which cannot be characterized via conventional two-stage methods. To this end, we reformulate the task as an offline constrained deep RL problem. \textbf{(ii)} To reduce the memory cost of mixed policy methods, we propose two model-agnostic methods, i.e., AIM-mean and AIM-greedy, which effectively reduce the number of stored individual policies to a constant. Our methods are general solutions to the critical issue of memory consumption in mixed policy methods. \textbf{(iii)} Empirical results demonstrate the superior performance and policy efficiency of the proposed method, which has been successfully deployed on a large-scale online marketing campaign with tens of millions of users and a more than one billion marketing budget.

\section{Related Work}
This section reviews prior works on marketing budget allocation.\\
\textbf{Classic methods}. Many classic methods use a two-step framework \cite{zhao2019unified,agarwal2014budget,xu2015smart,chen2021adversarial,liu2019graph,li2021large,yu2021joint}, i.e., a response prediction step and a decision making step. Response prediction models include DNN \cite{chen2021adversarial,zhao2019unified} or GNN \cite{liu2019graph,yu2021joint}, while the decision making step can utilize dual method \cite{zhong2015stock}, linear programming \cite{yu2021joint,chen2021adversarial}, bisection method \cite{zhao2019unified} or control based method \cite{xu2015smart}. These methods only optimize the immediate reward and fail to capture the long-term effect.

\noindent\textbf{Constrained RL}. Constrained Markov Decision Process (CMDP) \cite{altman1999constrained} is a commonly used framework to characterize the long-term effect. When the problem scale is small, CMDP can be directly solved via dynamic programming \cite{labbi2007optimizing} or linear programming  \cite{altman1999constrained}. In large-scale industrial scenarios where user transitions are very hard to model \cite{chen2021survey}, many methods learn optimal policy by directly interacting with the environment rather than modeling the whole dynamics of the environment. Though policy-based methods such as CPO \cite{achiam2017constrained} and RCPO \cite{tessler2018reward} have success in certain tasks, policy-based methods are rarely adopted in industrial-level usages \cite{afsar2021reinforcement}. Because the extremely large user and item spaces in industrial applications lead to large variance and low sample efficiency when learning from offline datasets \cite{zou2020pseudo}, value-based or actor-critic methods are more commonly used \cite{chen2021survey}. However, it has been shown that directly applying them may not converge \cite{altman1999constrained,abernethy2011blackwell}. Although mixed policy methods \cite{miryoosefi2019reinforcement,le2019batch,bohez2019value,zahavy2021reward} are proved to converge to the optimal policy, they conventionally need to store all historical policies. Hence, the memory cost is linear to the number of training steps, which is prohibitively expensive in large-scale scenarios. 

% \citet{xiao2019model} proposes a value-based method to solve constrained RL by penalizing the Q-value function with constraint dissatisfaction.

\section{Long-Term Effect in marketing budget allocation}

In online marketing campaigns, the core problem is to appropriately allocate budgets to incentivize desired user behaviors while satisfying certain budget constraints. The effectiveness of a marketing campaign is commonly measured by average DAU over different time windows.

As an example, we here discuss a typical scenario of marketing budget allocation, which was held by Alipay from Q4 2020 to Q2 2021 to improve the market share of its digital payment app. During the campaign, tens of millions of users participated and cash coupons worth more than a billion yuan were distributed and redeemed. Illustrated by Figure \ref{fig:caption}, they offer cash coupons to their users, who can redeem one cash coupon per day when paying with the app and get another one for the next day. The cash coupon is used to incentivize users to try out the digital payment app, and the users are expected to enjoy the convenience and establish a new paying habit. The performance of the campaign is measured by the weekly average DAU who make digital payments. With a slight abuse of notion, when discussing on the user level, we use average DAU to refer to the number of days the user makes payments, whose sum over all users yields the average DAU of the app.

Conventional approaches simplify the above long-term (i.e., one-week) decision making problem to an optimization task over short-term metrics \cite{yu2021joint,chen2021adversarial,liu2019graph}. These methods first estimate some short-term metrics, like the coupon redemption rate or the users' DAU in the coming day, under different coupon values. They then perform an optimization task to make the budget allocation \cite{agarwal2014budget,zhong2015stock}. In the following, via careful investigation on the real data, we show that the above methods actually oversimplify the marketing budget allocation task, neglecting the important long-term effect.

We first verify whether the simplification of substituting long-term metrics with short-term metrics is reasonable in marketing budget allocation tasks. We collect a dataset of 4.2 million samples with a random budget allocation strategy (data collection is discussed in Sec.\ref{data_collection}). Near the budget averages, we randomly assign different coupon amounts to the users and collects the users' DAU metrics over 1/3/7 days' time frames.

Given a fixed budget, the optimal budget allocation plan depends on sensitivity of the DAU metrics with respect to the changes in coupon values \cite{yu2021joint,liu2019graph,zhong2015stock}. To simplify the analysis, we set the coupon value to be 1 or 3 yuan, and measure the difference in the corresponding users' DAU. As a common practice in marketing budget allocation \cite{labbi2007optimizing}, we divide all users into an active group and an inactive group according to their historical DAU, and inspect their behaviors respectively. Intuitively, the optimal budget allocated to a group of users should be approximately proportional to their DAU changes \cite{yu2021joint}. We then compare the DAU changes after allocating this additional budget to each user. The result, as illustrated in Table \ref{table_ltv}, shows that the additional 2 yuan has a 32\% additional effect for inactive users over a 1-day window. However, the margin enlarges to a more significant 82\% after 7 days. This shows that if we only consider the short-term metrics, it is likely that we significantly underestimate the potential gain of allocating more budgets to inactive users, resulting in a non-optimal policy.

\begin{table}[tb]
  \caption{The DAU improvement of increasing the coupon from 1 to 3 yuan over different user groups \& time frames.}
  \vspace{-3mm}
  \label{table_ltv}
\begin{tabular}{ccc}
    \toprule
    \makecell{DAU change \\ over} & \makecell{Inactive \\ users} & \makecell{Active \\ users}\\% & Rate\\
    \midrule
    1 day & +0.15 & +0.11  \\%& 132\%\\
    3 days &  +0.54 & +0.31 \\%& 172\%\\
    7 days & +1.24 & +0.68 \\% & 182\%\\
  \bottomrule
\end{tabular}
\quad 
\begin{tabular}{c}
    \toprule
    \makecell{Diff. \\ (Inactive/Active - 1)}\\
    \midrule
    32\%\\
    72\%\\
    82\%\\
  \bottomrule
\end{tabular}
\vspace{-3mm}
\end{table}

\section{Offline Constrained RL for marketing budget allocation}

In this section, we formalize the marketing budget allocation problem as a \textit{offline constrained RL} task and review the mixed policy approach \cite{le2019batch,miryoosefi2019reinforcement} to solve it.

\subsection{Problem Formulation}
A well-established framework for studying constrained RL is the Constrained Markov decision process (CMDP) \cite{altman1999constrained}, which can be identified by a tuple $\{\sS,\sA,P,r,\vc\}$. Over an infinite time horizon, each time at $t=1,2,...$, the agent observes a state $s_t\in \sS$ among the state space and chooses an action $a_t\in\sA$ among the action space. Given the current state and the action chosen, the state at the next observation evolves to state $s_{t+1} \in \sS$ with probability $P(s_{t+1} | s_t, a_t)$, and the agent receives an immediate reward $r_t \in \R$. In particular, for the marketing budget allocation problem, the corresponding action $a_t$ consumes the limited budget that constrains the usage of resources. Therefore, we assume that action $a_t$ suffers an $m$-dimensional immediate cost $\vc_t \in \R^m$, where $m$ is the number of constraints. The decisions are made according to a policy $\pi$. At each state, a policy is a distribution over action space. 

% \textcolor{red}{1. authors did not clearly state what is the immediate reward and cost in their framework. I had to guess.}

In the following, we consider two types of policies. Deterministic policy chooses a single action deterministically for any state, and can be identified by the mapping $\pi: \mathcal{S} \mapsto \mathcal{A}$. The set of all deterministic policies is denoted by $\Pi$. To improve the convergence of value-based methods, it is helpful to consider \textit{mixed} deterministic policy $\Delta(\Pi)$ \cite{le2019batch,bohez2019value,miryoosefi2019reinforcement,zahavy2021reward}, where $\mu \in \Delta(\Pi)$ is a distribution over $\Pi$. To execute a policy $\mu \in \Delta(\Pi)$, we first select a policy $\pi\in\Pi$ according to $\pi\sim\mu$ and then execute $\pi$ for the entire episode. 

For any discount factor $\gamma \in [0,1)$ and any policy $\pi\in\Delta(\Pi)$, the expectations of the discounted reward $J_r(\pi)$ and the discounted cost $J_\vc(\pi)$ are defined as 
\begin{align*}
    J_r (\pi) := \E\left[\sum_{t=1}^{\infty} \gamma^{t-1} r_t\right],
    \quad
    J_\vc (\pi) := \E\left[\sum_{t=1}^{\infty} \gamma^{t-1} \vc_t\right],
\end{align*}
where the expectation is over the stochastic process. To simplify notions, we define a policy's \textit{measurement vector} as the $(m+1)$-dimensional vector containing the expectations of the discounted reward  $J_r(\pi)$ and the discounted cost $J_\vc(\pi)$ ,
\begin{align*}
\vx(\pi) := [J_r(\pi), J_\vc(\pi)] \quad \text{(Measurement vector)}.
\end{align*}

In offline constrained RL, we have a pre-collected dataset $D := \{(s_i, a_i, s_i', r_i, \vc_i)\}_{i=1}^n$ containing transitions generated from (possibly multiple) behavior policies, jointly denoted as $\pi_B$. The goal in offline constrained RL is to find an \textit{optimal policy}, denoted as $\mu^*$, that maximizes the reward subject to the constraints
\begin{align}
    \max_{\mu\in\Delta(\Pi)} J_r(\mu) \quad \text{subject to } \quad J_\vc(\mu) \preceq \tau \label{ocrl}
\end{align}
where $\tau\in\R^m$ is vector of known constants. For an optimal policy $\mu^*$, $J_r(\mu^*)$ is called the optimal value of the problem. To simplify the discussion, for any $\mu$ not satisfying the constraints, let $J_r(\mu) := -\infty$.

Without the constraints, learning optimal policy from a pre-collected dataset is a well-studied problem. For such problems, many effective methods rely on Q-learning to learn in an off-policy style and find optimal deterministic policies efficiently \cite{sutton2018reinforcement}. However, different from the unconstrained case, for constrained RL, none of the optimal policies may be deterministic. We illustrate this with the following example.

\begin{example} \label{example1}
Consider a CMDP with a single state, and two actions, where the first action has $r=0, c=0$ and the second action with $r=1, c=1$. The process terminates whenever an action is taken. For any $\tau\in(0,1)$, to maximize the reward subject to $J_c\le \tau$ requires randomization among the two actions with the second action being chosen with probability $\tau$. In this case, the optimal policy exists and is not deterministic.
\end{example}

Therefore to solve offline constrained RL, it is common to consider using mixed deterministic policy. 
% Though methods converges to the optimal policy, approximates the exact solution requires storing infinitely many policies, which compromises the practical feasibility of such method, and is only of theoretical interest. In the next section, extending previous work, we propose a novel algorithm, which converges to the optimal policy which storing at most $m+1$ policies throughout the learning process. We further show that the $m+1$ constant is optimal.

% \section{Approach, Algorithm and Analysis}

\subsection{Relax to an Unconstrained Problem}

We now show that the offline constrained RL problem (\ref{ocrl}) can be reformulated to an \textit{equilibrium} problem (\ref{equilibirum_finding}), which can then be solved using online learning techniques.

We first relax the constrained problem (\ref{ocrl}) into an equivalent unconstrained optimization problem, where the value function is penalized by the constraint violation. Using a Lagrangian multiplier $\vlambda\in\R^m_+$, we construct the Lagrangian
\begin{align*}
    L(\mu, \vlambda) := J_r(\mu) - \vlambda^T (J_\vc(\mu) - \vtau).
\end{align*}
We show that solving the optimal value of the original constrained problem (\ref{ocrl}) is equivalent to solving the equilibrium or saddle point of this unconstrained problem
\begin{align}
\max_{\mu\in\Delta(\Pi)}\min_{\vlambda \in \R^m_+} L(\mu, \vlambda). \label{equilibirum_finding}
\end{align}
By a standard Lagrangian argument, $\forall \mu\in\Delta(\Pi)$ that violates certain constraints, by setting the corresponding dimensions of $\vlambda$ to $\infty$, and $0$ otherwise, then $L(\mu, \vlambda) = -\infty$. On the other hand, for any feasible policy $\mu$, since all elements in $J_\vc(\mu) - \tau$ are non-positive, the inner minimization is achieved when $\vlambda=\vzero$. The outer maximization is then realized by certain feasible policies that maximize the discounted reward $J_r(\mu)$. Hence, we conclude the equivalence between the constrained task (\ref{ocrl}) and the unconstrained one (\ref{equilibirum_finding}).

\begin{algorithm}[tb]
\caption{Original mixed policy method}
\label{alg:algorithm_1}
\begin{flushleft}
\textbf{Input}: no-regret online learning algorithm: $\mathtt{Learner}(\cdot)$; offline RL algorithm that yields a policy: $\mathtt{Best\_response}(\vlambda_{t})$\\
\textbf{Initialization}: $\vlambda_0 \in \R^m_+, \pi_0 \in \Pi$.
\end{flushleft}
\begin{algorithmic}[1] %[1] enables line numbers
\FOR{$t=1,2,...,T$}
    \STATE $\vlambda_t \gets \mathtt{Learner}(L(\pi_0, \cdot), \dots, L(\pi_{t-1}, \cdot))$
    \STATE $\pi_t \gets \mathtt{Best\_response}(\vlambda_{t})$
    \STATE $\mu \gets \{ \pi_i | i=1,2,\dots,t, \mu(\pi_i) = 1/t\}$ \COMMENT{Mixed policy that uniformly randomly plays $\pi_1, \dots, \pi_t$} \label{maintain_mixed_policy}
\ENDFOR
\STATE \textbf{return} $(\mu, \hat{\vlambda}): \vx(\mu) = \frac{1}{T} \sum_{t=1}^T \vx(\pi_t), \hat{\vlambda} =\frac{1}{T} \sum_{t=1}^T \vlambda_t $
\end{algorithmic}
\end{algorithm}

% where best-response($\lambda_t) = \arg\min_{\pi\in\Pi} L(\pi, \lambda_t)$, and $L(\pi, \lambda) = C(\pi) + \lambda^T(G(\pi-\tau))$ for $\lambda \in \mathbb{R}^m_+$. Online learning subroutine: similar to (Agarwal et al. 2018), they use Exponentiated Gradient (EG) (lower regret bound than OGD). Forces $||\lambda||_1\le B$ (bounded $\lambda$). Augment $\lambda$ into $m+1$ dimension by appending $B-||\lambda||_1$ and augment the constraint cost vector g by appending 0. 

\subsection{Online Learning Techniques}

We then take a game-theoretic perspective and solve the equilibrium finding problem (\ref{equilibirum_finding}) with online learning. From a game-theoretic perspective, the equilibrium finding problem (\ref{equilibirum_finding}) can be viewed as a two-player zero-sum game. The $\mu$ player plays against $\vlambda$ player, where $L(\mu, \vlambda)$ is the utility gained by $\mu$ player and the loss suffered by the $\vlambda$ player. The equilibrium of this game can be solved by the general technique of \cite{freund1999adaptive} by repetitively playing a \textit{no-regret} online learning algorithm against a \textit{best response} learner. 

We take a brief review of the online learning framework, where the core idea is the existence of no-regret learners \cite{shalev2011online}. For a learner who makes a sequence of decisions, the no-regret property means the cumulative loss by the current learning algorithm converges to the best-fixed strategy in the hindsight sublinearly. That is, after making $T$ decisions,
\begin{align*}
    \mathtt{Regret}_T := \sum_{t=1}^T L(\mu_t, \vlambda_t) - \min_{\vlambda \in \R^m_+} \sum_{t=1}^T L(\mu_t, \vlambda) =  o(T)
\end{align*}
where $\vlambda_t = \mathtt{Learner}(L(\mu_1, \cdot), \dots, L(\mu_{t-1}, \cdot))$ is generated by the online learning algorithm given the loss function of all previous plays. We also consider learners that are \textit{prescient}, i.e. that can choose $\vlambda_t$ with the knowledge of $L(\mu_t, \cdot)$ the loss function up to and including time $t$.

Perhaps the most trivial strategy in a prescient setting is simply to play the best choice for the current loss function. The best response strategy assumes the knowledge of the other player's play, and simply maximizes utility for the current loss function. For the $\mu$ player, this can be defined as 
\begin{align*}
\mathtt{Best\_response}(\vlambda_t) := \argmax_{\mu \in \Delta(\Pi)} L(\cdot, \vlambda_t)
\end{align*}

\textit{Online Gradient Descent} (OGD) \cite{zinkevich2003online} is a typical online learning algorithm, defined as follows
\begin{align}
\vlambda_{t+1} = \mathtt{OGD}(\vlambda_t, L) := \max(\vzero, \vlambda_t - \eta_t \nabla L). \label{ogd}
\end{align}
By setting appropriate learning rate, it has no-regret. Usually, the learning rate can be set adaptively with $\eta_t := 1/\sqrt{t}$, then the regret bound is $O(1/\sqrt{T})$ \cite{zinkevich2003online}.

% "An example of such (no-regret online learning algorithm) is \textit{online gradient descent (OGD)} of \cite{zinkevich} (see Appendix \ref{}). If the Euclidean diameter of $\Lambda$ is at most  $D$, and $||\nabla L_t(\lambda) \le G||$  for any $t$ and $\lambda \in \Lambda$, then the regret of the OGD is at most $DG\sqrt{T}$. "

\subsection{Solve the Unconstrained Problem}
The method of Freund and Schapire  \shortcite{freund1999adaptive} to solve the equilibrium finding problem can be summarized as Algorithm \ref{alg:algorithm_1}, where the $\mathtt{Learner}$ can be, for example, the OGD defined in (\ref{ogd}). As stated in the following proposition, the average of players' decisions converges to the equilibrium point.

\begin{proposition} \label{meta_algo_thm}
For Algorithm \ref{alg:algorithm_1}, let $\mathtt{Regret}_T$ be the regret of the online learning algorithm used by $\vlambda$-player, then after $T$-round of iterations, Algorithm \ref{alg:algorithm_1} returns $(\mu, \hat{\vlambda})$ with duality gap bounded by
\begin{align}
    \max_{\mu \in \Delta(\Pi)} L(\mu, \hat{\vlambda}) - \min_{\vlambda \in \R^m_+} L(\mu, \vlambda)  \le \frac{\mathtt{Regret}_T}{T}
\end{align}
\end{proposition}
\begin{proof}
For any given $\hat{\vlambda}$, maximize $L(\cdot, \hat{\vlambda})$ is equivalent to an ordinal unconstrained RL problem, where at each step a penalized reward $r' := r - \vlambda^T(\vc - \tau)$ is given.  Therefore deterministic policies that maximize the $r'$, also maximizes $L(\cdot, \hat{\vlambda})$ and we have 
\begin{align*}
    \max_{\mu \in \Delta(\Pi)} L(\mu, \hat{\vlambda})  = \max_{\pi \in \Pi} L(\pi, \hat{\vlambda}).
\end{align*}
Note that this is critical to ensure the convergence of RL algorithms that yield deterministic policies. Since $L(\cdot, \vlambda)$ is linear to $\vlambda$, and $\pi_t$ are $\mathtt{Best\_response}(\vlambda_{t})$ policies,
\begin{align*}
     \max_{\pi \in \Pi} L(\pi, \hat{\vlambda}) = \max_{\pi \in \Pi}\frac{1}{T} \sum_{t=1}^T L(\pi, \vlambda_t) 
    \le \frac{1}{T} \sum_{t=1}^T L(\pi_t, \vlambda_t)
    % &= \frac{1}{T} \min_{\vlambda\in\R^m_+} \sum_{t=1}^T L(\pi_t, \vlambda) +  \frac{\mathtt{Regret}_T}{T}\\
    % \max_{\mu \in \Delta(\Pi)} L(\mu, \hat{\vlambda}) - \min_{\vlambda \in \R^m_+} L(\mu, \vlambda)  \le \frac{\mathtt{Regret}_T}{T}
\end{align*}
Then the claim follows from the definition of $\mathtt{Regret}_T$.
\end{proof}
% \textcolor{red}{expain the average, single best. 5– the statement of the theorem 4.1 and 4.2 is not clear nor self-contained. For example, what is the running average, inputs policies, and single best policy?
When running Algorithm \ref{alg:algorithm_1} in practice, $\hat{\vlambda}$ can be easily maintained in place as a vector, and updated via $\hat{\vlambda} = \hat{\vlambda} \cdot (t-1)/t + \vlambda_{t} /t$ at each round. However, it is much more difficult to maintain the mixed policy $\mu$ in our scenario where deep neural networks are typically used as function approximators. Since neural networks are not linear, to recover the mixed policy, it requires to store the parameters of all individual policies $\{\pi_1, \dots, \pi_T\}$. The memory cost is prohibitively expensive since the neural networks can be considerably large.

To overcome this problem, in the following section, we introduce a novel notion of Affinely Independent Mixed (AIM) policy, which reformulates any mixed policy based on only constantly many individual policies, which successfully reduces the memory cost of mixed policy methods to a constant.

\section{AIM-based methods}

In this section, we first propose the notion of AIM policy, which can be recovered via randomly sampling from a set of individual policies that are affinely independent from each other in the measurement space. Then we propose the AIM-mean method to equivalently transform the mixed policy in Algorithm \ref{alg:algorithm_1} into an AIM policy, which maintains a constant number of individual policies throughout the training process while still ensuring convergence. We finally propose a variant called AIM-greedy by discarding the undesirable individual policies in AIM-mean, which outperforms the heuristic method using any single policy \cite{le2019batch,xiao2019model} empirically and is favorable for practical usage.

Note that although our AIM method is motivated by the marketing budget allocation problem, it is actually suitable to a large range of scenarios using deep RL with mixed policies \cite{zahavy2021reward}, such as apprenticeship learning \cite{zahavy2020apprenticeship,abbeel2004apprenticeship} and pure exploration \cite{hazan2019provably}. Our AIM method is a general approach that fundamentally solves the issue of storing all historical policies in mixed policy methods, which can be used in these deep RL tasks as well.

\subsection{Affinely Independent Mixed (AIM) Policy}

We first introduce some notations and definitions regarding mixed policy. For any mixed policy $\mu\in\Delta(\Pi)$, we define its active policy set $S_p(\mu)$ as the set of all deterministic policies with a non-zero probability in $\mu$. i.e.,\footnote{We here slightly abuse $\mu(\pi)\in[0, 1]$ to denote the probability of any deterministic policy $\pi\in\Pi$ in the mixed policy $\mu$.}
\begin{align*}
S_p(\mu) := \{\pi | \pi\in\Pi, \mu(\pi)>0\}.
\end{align*}
We also define its active measurement set $S_x(\mu)$ as the set of the measurement vectors of all these active policies, i.e.,
\begin{align*}
S_\vx(\mu) := \{\vx(\pi) | \pi\in S_p(\mu)\}.
\end{align*}

%Therefore, by carefully selecting the AIM policies, any mixed policy can be equivalently implemented by randomly sampling from these finite policies with some proper weights. Since a $(m+1)$-dimensional space (i.e., a reward and $m$ costs) can be spanned by $m+2$ affinely independent vectors, intuitively we can reduce the number of stored individual policies to a constant. 

%To deal with the scale problem, we introduce the Affinely Independent Mixed (AIM) policy. It maintains a mixed policy in a policy-efficient manner therefore achieving memory efficiency. The idea is to select a subset of policies whose measurement vectors are affinely independent and maintain a proper weight, to express the target mixed policy. 

Recall that the measurement vector of $\mu$ is given as
\begin{align*}
\vx (\mu) = \E_{\pi\sim\mu}\left[\vx(\pi)\right] = \sum_{\pi\in\Pi} \vx(\pi) \mu(\pi),
\end{align*}
Suppose the active policy set is finite, i.e., $S_p(\mu)=\{\pi_1,...,\pi_k\}$, then we equivalently have
\begin{align*}
\vx(\mu) = [\vx(\pi_1), ..., \vx(\pi_k)] \begin{bmatrix}
       \mu(\pi_1) \\
       \vdots \\
       \mu(\pi_k) 
     \end{bmatrix},
\end{align*}
which is a convex combination of its active measurement vectors $\vx(\pi_1),...,\vx(\pi_k)$ in $S_\vx(\mu)$.

We call $\mu$ an AIM policy, if its active measurement vectors are affinely independent of each other. Intuitively, the AIM policy characterizes a family of mixed policies with a relatively small active policy set. Specifically, if some mixed policy $\mu'\in\Delta(\Pi)$ is not an AIM policy, namely the elements in $S_\vx(\mu')$ are affinely dependent, we can pick several affinely independent elements such that $\vx(\mu')$ lies in the convex hull of these elements. In this case, by adjusting the mixing weights accordingly, the mixed policy $\mu'$ can be recovered with a smaller active policy set.

To formalize the above reduction, notice that for an AIM policy $\mu$, its active measurement vectors are affinely independent and form a \textit{simplex} in $\R^{m+1}$. For any policy $\mu'$ whose measurement vector $\vx(\mu')$ lies in the affine hull $\aff(S_\vx(\mu))$, its \textit{barycentric coordinate} $\valpha\in\R^{k}$ can be solved via
\begin{align} \label{b_coord}
\vx(\mu') = [\vx(\pi_1)\,\dots\,\vx(\pi_k)]\valpha, \quad\text{s.t.} \sum_{i} \alpha_i = 1.
\end{align}
If $\alpha_i\ge 0$ for any $i$, then $\vx(\mu')$ lies inside the convex hull $\conv(S_\vx(\mu))$. In this case, we can achieve the same measurement vector $\vx(\mu')$ of $\mu'$ by randomly choosing the policy $\pi_i\in S_p(\mu)$ with probability $\alpha_i$. In this way, we only need to store the policies in $S_p(\mu)$, which might be smaller than $S_p(\mu')$. 

In summary, any mixed policy can be equivalently realized by randomly sampling with proper weights from several deterministic policies that are affinely independent in the measurement space $\mathbb R^{m+1}$. Since $\mathbb R^{m+1}$ is the affine span of any $m+2$ affinely independent vectors, intuitively we can reduce the number of the stored individual policies to at most $m+2$. In the following, we devise two algorithms that realize the AIM policy. 

\subsection{AIM-mean Algorithm}
We first propose AIM-mean (Algorithm \ref{alg:aim_mean}), which realizes the original mixed policy method (Algorithm \ref{alg:algorithm_1}) via AIM policies. At each round $t$, AIM-mean outputs an AIM policy whose measurement vector is equal to the average measurement of all historical policies, i.e.,
\begin{align*}
    \vx(\mu_t) = \frac{1}{t}\sum_{i=1}^t \vx(\pi_i).
\end{align*}
Therefore, AIM-mean evolves exactly the same as the original method, which ensures convergence to the optimal policy, while requiring to store no more than $m+2$ individual policies.

The core step of AIM-mean is the maintenance of the active policy set of the output mixed policy, which we now explicate in detail. At the beginning of each round $t$, the algorithm receives a new policy $\pi_t$ generated by the original algorithm, whose measurement vector takes $\vx(\pi_t) \in\R^{m+1}$. After receiving the policy, we first calculate the target vector $\vx_t$ (i.e., the average measurement vector) at the current round, then decide whether we should alter the active set:

\begin{algorithm}[tb]
\caption{AIM-mean}
\label{alg:aim_mean}
\begin{flushleft}
\textbf{Input}: AIM-mean policy $\mu_{t-1}$, any $\pi_t$ with $x(\pi_t) \in \R^{m+1}$\\
% \textbf{Parameter}: Optional list of parameters\\
\textbf{Output}: AIM-mean policy $\mu_t$ s.t. $\vx(\mu_t) = \frac{1}{t}\sum_{i=1}^t \vx(\pi_i)$
\end{flushleft}
\begin{algorithmic}[1] %[1] enables line numbers
%  TODO: not correct, may not maintain the Affinely independent property
\STATE \label{line1} $\vx_t \gets \vx(\mu_{t-1}) * (t-1)/t + \vx(\pi_t)/t$ \COMMENT{Target value}
\IF {$\vx_t$ not in affine hull $\aff (S_\vx(\mu_{t-1}))$}
\STATE $S_\vx \gets S_\vx(\mu_{t-1}) \cup \{\vx(\pi_t)\}$ 
\ELSIF{$\vx_t$  in convex hull $\conv (S_\vx(\mu_{t-1}))$}
\STATE $S_\vx \gets S_\vx(\mu_{t-1})$
\ELSE 
% \STATE $S_\vx \gets S_\vx(\mu_{t-1}), \vx_{t-1} = \vx(\mu_{t-1})$, not\_added = True
% \WHILE{$\vx_t$ not in convex hull $\conv(S_\vx)$}
% \STATE $S_\vx, \vx_{t-1} \gets \mathtt{RemoveOneVertex}(S_\vx, \vx_{t-1}, \vx_t) $
% \IF{not\_added}
% \STATE $S_\vx \gets S \cup \{\vx(\pi_t)\}$, not\_added = False
% \ENDIF
% \ENDWHILE

\STATE $S_\vx \gets S_\vx(\mu_{t-1}), \vx_{t-1} = \vx(\mu_{t-1})$
\STATE $S_\vx, \vx_{t-1} \gets \mathtt{RemoveOneVertex}(S_\vx, \vx_{t-1}, \vx_t) $
\STATE $S_\vx \gets S \cup \{\vx(\pi_t)\}$
% \ELSE 
%
\ENDIF
\REPEAT
\STATE $S_\vx, \vx_{t-1} \gets \mathtt{RemoveOneVertex}(S_\vx, \vx_{t-1}, \vx_t) $
\UNTIL {no more vertex can be removed}
\STATE $\valpha \gets $ barycentric coordinate of $\vx_t$ w.r.t. $S_\vx$
\STATE \textbf{return} $\mu_t \gets S_\vx \alpha$
\end{algorithmic}
\end{algorithm}

% \textcolor{red}{2. how did the authors guarantee its policy's active measurement vectors are affinely independent?
% }

(i) If $\vx_t\notin\aff(S_\vx(\mu_{t-1}))$, this implies that the size of $S_\vx(\mu_{t-1})$ is less than $m+2$ (otherwise we have $\aff(S_\vx(\mu_{t-1}))=\mathbb R^{m+1}$). In this case, $\vx(\pi_t)$ is affinely independent of any vector in the current active set $S_\vx(\mu_{t-1})$, and we add $\vx(\pi_t)$ to the active set. 

(ii) If $\vx_t\in\conv(S_\vx(\mu_{t-1}))$, then we maintain the active set as $S_\vx(\mu_{t-1})$. Now we just need to recalculate the mixing weights of the elements in $S_\vx(\mu_{t-1})$ regarding $\vx_t$ via (\ref{b_coord}).

(iii) If $\vx_t\in\aff(S_\vx(\mu_{t-1}))$ and $x_t\notin\conv(S_\vx(\mu_{t-1}))$, then we need to alter the elements in active set and express $\vx_t$ as a convex combination. Intuitively, we can replace one element in $S_\vx(\mu_{t-1})$ by $\vx(\pi_t)$ so that the simplex formed by the new active set contains $\vx_t$. Notice that since $\vx(\mu_{t-1})$ and $\vx_t$ lie inside and outside of the original simplex respectively, the segment joining the two points must cross the boundary of the original simplex. Therefore, we can find the specific facet where the intersection locates at, remove any vertex in $S_\vx(\mu_{t-1})$ that is not on this facet, and add the new policy to the active set. 

The above procedure of removing vertex is summarized as the $\mathtt{RemoveOneVertex}(S_\vx, \vx_{t-1}, \vx_t)$ subroutine, which takes three arguments, namely the vertices of the simplex $S_\vx$, previous average measurement vector $\vx_{t-1}$, and the target vector $\vx_t$. $\mathtt{RemoveOneVertex}$ first finds the intersection of the above segment and the facet of the simplex, then replaces a random vertex that is not on the facet with the new policy $x(\pi_t)$ at the current round. Specifically, assume  $\valpha_{t-1}$ and $\valpha_t$ as the barycentric coordinates of $\vx_{t-1}$ and $\vx_{t}$ with respect to $S_\vx$. Denote $\valpha_t=(\alpha_{t,1},...,\alpha_{t,m+1})$. Since $\vx_t\notin\conv(S_\vx)$, some entries of $\valpha_t$ must be non-positive. Hence, we can set
\begin{align} \label{move_to_boundary}
    \theta \gets \arg\min_{i: \alpha_{t,i}\le 0}   \frac{\alpha_{t-1,i}}{\alpha_{t-1,i} - \alpha_{t,i}}.
\end{align}
Then $\valpha_t \gets \theta \valpha_t + (1-\theta) \valpha_{t-1}$ is the point on the facet of the simplex. Now we just need to remove any vertex in $S_\vx$ whose weight corresponding to $\alpha_t$ is non-positive, and update $\vx_{t-1} \gets \theta \vx_t + (1-\theta) \vx_{t-1}$ accordingly. 

After updating the active set $S_\vx$ and the mixing weights $\valpha_t$, we repeatedly call $\mathtt{RemoveOneVertex}$ to remove any vertex that does not contribute to $\vx_t$. This further reduces the number of the stored individual policies without affecting the output mixed policy $\mu_t$.

It can be easily checked that at the end of each round, $S_\vx$ is an affinely independent set and $\vx_t\in\conv(S_\vx)$. This validates the correctness of AIM-mean, as summarized in the following.

\begin{corollary} \label{thm4_2}
AIM-mean maintains the running average of the measurement vectors of all input policies as an AIM policy.
\end{corollary}

\begin{table*}[t] \label{table_compare}
  \caption{Performance comparison of different methods on Alipay dataset.}
  \vspace{-3mm}
  \label{table_compare}
  \begin{center}
  \begin{tabular}{ccccccccc} 
  \toprule % \multirow{2}{*}{Method}
   & \multicolumn{2}{c}{\multirow{2}{*}{Method}}  & \multicolumn{3}{c}{\textbf{Simplified Dataset}} & \multicolumn{3}{c}{\textbf{Full Dataset}}\\ \cline{4-9}
    & & & 1d DAU & 3d DAU & 7d DAU & 1d DAU &  3d DAU & 7d DAU \\ \midrule 
    \multirow{2}{*}{Classic} 
    & DeepFM & - & 0.5997	& 2.2032	& 4.7806	& 0.6396	& 2.3232	& 5.2114 \\
    & DIN & - & \textbf{0.6010}	& 2.2070	& 4.7854	& \textbf{0.6468}	& 2.3306	& 5.2237\\  \midrule
    \multirow{6}{*}{Off-policy RL} 
    %     & \multirow{3}{*}{\shortstack[c]{DDPG-att\\(EDRR)}} & Single-best &&&&&&\\
    % &   & AIM-mean &&&&&&\\
    % &   & AIM-greedy &&&&&&\\ \cline{2-9}
    & \multirow{3}{*}{\shortstack[c]{DDQN}} & Single-best & 0.5272	& 2.3597	& 4.9418	& 0.5981	& 2.5014	& 5.6306\\
    &   & AIM-mean & 0.5172	& 2.3433	& 4.9360	& 0.5779	& 2.4819	& 5.5759\\
    &   & AIM-greedy & 0.5339	& 2.3640	& 4.9732	& 0.5998	& 2.5213	& 5.6361 \\ \cline{2-9}
    & \multirow{3}{*}{\shortstack[c]{EDRR}} & Single-best & 0.5741	& 2.4003	& 4.9620	& 0.6156	& 2.5243	& 5.6390 \\
    &   & AIM-mean & 0.5572	& 2.3849	& 4.9360	& 0.6045	& 2.5144	& 5.6255 \\
    &   & AIM-greedy & 0.5893	& 2.4193	& 5.1153	& 0.6163	& 2.5253	& 5.6432 \\
    \midrule
    \multirow{3}{*}{Offline RL}
    %     & \multirow{3}{*}{BCQ-PG} & Single-best &&&&&&\\
    % &   & AIM-mean &&&&&&\\
    % &   & AIM-greedy &&&&&&\\ \cline{2-9}
    & \multirow{3}{*}{BCQ} & Single-best & 0.5770	& 2.4068	& 5.0608	& 0.6150	& 2.5318	& 5.6901\\
    &   & AIM-mean & 0.5613	& 2.4016	 & 5.0461	& 0.6159	& 2.5310	& 5.6685\\
    &   & AIM-greedy & 0.5846	& \textbf{2.4266}	& \textbf{5.1911}	& 0.6205	& \textbf{2.5454}	& \textbf{5.7020} \\
    \bottomrule
\end{tabular}
\end{center}
\end{table*}

\subsection{AIM-greedy Algorithm}

\begin{algorithm}[tb]
\caption{AIM-greedy}
\label{alg:aim_greedy}
\begin{flushleft}
\textbf{Input}: AIM policy $\mu_{t-1}$, any $\pi_t$ with $x(\pi_t) \in \R^{m+1}$\\
% \textbf{Parameter}: Optional list of parameters\\
\textbf{Output}: AIM policy $\mu_t$
\end{flushleft}
\begin{algorithmic}[1] %[1] enables line numbers
\STATE Replace line \ref{line1} in AIM-mean by choosing $\vx_t\in[\vx(\mu_{t-1}), \vx(\pi_t)]$ that maximizes the reward at the premise of satisfying constraints:
%performing a line search on segment ($\vx(\mu_{t-1}), \vx(\pi_t)$) for $\vx_t$ closest to constraint satisfaction and then maximizes rewards.
%\STATE $\vx_t \gets \arg\max_{\vx\in[\vx(\mu_{t-1}), \vx(\pi_t)]}  \min_{\vlambda \in \R^m_+} L(\vx, \vlambda).$ 
\STATE $\vx_t \gets \arg\max_{\vx\in[\vx(\mu_{t-1}), \vx(\pi_t)]}  \min_{\vlambda \in \R^m_+} L(\vx, \vlambda).$ 
\end{algorithmic}
\end{algorithm}

% To improve the performance, we consider removing "bad" policies and propose the AIM-greedy method. Though this method may not converge, it is shown to always outperform the previous heuristic method and improve the practical performance significantly in our experiment.

Recall that classic mixed policy methods randomly select an individual policy from all historical policies via uniform sampling. There is another heuristic method \cite{le2019batch} that selects the \textit{single-best} policy among all individual policies, which improves the empirical performance despite lack of convergence guarantee. To improve the practical performance, we propose a novel variant called AIM-greedy by removing "bad" active policies in the mixed policy. Although it may not converge, AIM-greedy consistently outperforms the previous heuristic method in experiment, hence it is favorable for practical usage. 

As described in Algorithm \ref{alg:aim_greedy}, AIM basically follows the procedure of AIM-mean to ensure the AIM property, except that it uses a more radical target vector $\vx_t$ than AIM-mean. 

To decide the target vector, AIM-greedy performs a greedy line search on the segment between the previous target vector $\vx_{t-1}=\vx(\mu_{t-1})$ and the new policy $\vx(\pi_t)$, and selects a target vector $\vx_t$ that maximizes the reward at the premise of least constraint violation. Specifically, (i) if both $\vx_{t-1}$ and $\vx(\pi_t)$ satisfy the constraints, we set $\vx_t$ to the policy with a higher reward; (ii) if neither satisfies the constraints, we choose the policy whose Euclidean distance to the target constraints is smaller; (iii) If one satisfies while the other does not, and the infeasible policy has the higher reward, we move $\vx_t$ to the boundary of the constraints (i.e., line 6 in Algorithm \ref{aim_greedy}).

%Taking Example \ref{example1} for illustration, for two actions, the one that yields a higher reward violates the constraint. However, the one that satisfies the constraint has a lower reward. Using AIM-greedy, we maintain a mixed policy that takes exactly $\tau$ cost. 

In fact, AIM-greedy generalizes the idea of selecting the single best policy and is guaranteed to perform no worse than any single best policy. 

\begin{theorem} \label{aim_greedy}
The performance of the AIM-greedy algorithm is no worse than using any single best policy.
\end{theorem}
\begin{proof} 
At each round, AIM-greedy conducts a linear search on $\vx_t\in[\vx(\pi_t), \vx(\mu_{t-1})]$ to maximize $\min_{\vlambda \in \R^m_+} L(\vx_t, \vlambda)$, whereas the single best policy selects from $\vx_t\in\{\vx(\pi_t), \vx(\pi_{t-1})\}$. Using induction, since this equation is linear to $\vx(\mu_t)$, by the induction hypothesis, a line search on segment $(\vx(\pi_t), \vx(\mu_{t-1}))$  always performs no worse than selecting single best policy.
\end{proof}

\subsection{Policy efficiency}

We then discuss the optimal policy efficiency which shows that our AIM policy achieves optimal policy efficiency.

\begin{theorem} \label{storage}
For a constrained RL problem with $m$-dimensional constraint vector, a mixed policy needs to randomize among at least $m+1$ individual deterministic policies to ensure convergence.
\end{theorem}
\begin{proof} 
We give constructive proof. Consider a CMDP with a single state and $m+1$ actions. The first action has reward $r=0$, and cost $\vc=\vzero\in\R^m$. The rest actions all have reward $r=1$, and $\vc=\ve_i$ is the unit vector of the i-th dimension, and the episode terminates after any action is taken. Let $\tau = 1/(1+m) \vone$, the all one vector times $1/m$. Then it is clear that maximizing reward subject to these constraints requires randomizing among all $m+1$ actions uniformly. When deterministic policies are used, it requires randomizing among at least $m+1$ policies. 
\end{proof}

% \begin{corollary}
Since the affinely independent properties ensure the AIM-type policies store no more than $m+2$ policies throughout the learning process, and hence the storage complexity is optimal to ensure convergence when learning from off-policy samples. 
% \end{corollary}

\section{Experiments}
\subsection{Dataset} \label{data_collection}

During the marketing campaign, Alipay uses an online A/B testing system to perform user-level randomization; for a small percentage of users, a random budget allocation policy is used to assign coupon values. After running this randomized trial for a week in Dec 2020, 4.2 million samples are collected. Since the random strategy is likely to be non-optimal, given the average coupon amount of about 2.5 yuan, the collection of these samples is indeed expensive. Each sample has about one thousand raw features, which include timestamp, age, gender, city, and other behavior features. After one-hot encoding, all samples take around one hundred thousand different values. The label of each sample is the corresponding user's DAU over the next 1/3/7 days.

The full dataset takes about 56 gigabytes on disk, and our deployed model is implemented with the company's private distributed model training infrastructure that supports high-dimensional embedding lookup tables. To simplify the tuning process, we also create a simplified data set, which contains only 2 user features, 1) whether the user participated in the previous day, which is a boolean value, and 2) the activeness level of the user. The results on both datasets are qualitatively similar and are both reported.

\subsection{Evaluation Setting}

\noindent{\textbf{The baselines.}} We first compare our proposed method with classic methods that optimize short-term metrics (i.e. 1d DAU). For classic methods, we adopt the widely used \textit{DeepFM} \cite{guo2017deepfm} and \textit{DIN} \cite{zhou2018deep}. We then compare our proposed AIM-mean and AIM-greedy methods with the heuristic single-best method \cite{le2019batch}. Since these methods are model-agnostic, we implement them using the following RL methods. (i) \textit{DDQN} uses double Q techniques \cite{van2015deep,fujimoto2018addressing} on the vanilla DQN \cite{mnih2013playing,liu2018deep} to avoid overestimation. We set the weight for the smaller Q network to 0.8, and the higher one to 0.2. (ii) \textit{EDRR} \cite{liu2020end} is a state-of-the-art method in RL-based recommendations, which augments the aforementioned DDQN methods with a supervised task to stabilize the training of the embedding layer. (iii) \textit{BCQ} \cite{fujimoto2019off} adds a behavior cloning task to encourage the learned policy to stay close to the behavior policy that is being used to generate the offline dataset. We add the imitation task to the EDRR model and set the weight of behavior cloning to be $1.0$.

\noindent{\textbf{Training setting.}}
For each dataset, we divide 90\% of the data as the training set and leave the rest as the evaluation set. The division is performed by using a hashing function on the user id, to ensure that all interactions of a user either are in the training set or the evaluation set. Each method is trained for 2 epochs. The discount rate is set to be $0.8$, and the batch size is $512$. We utilize Adam optimizer for all the methods. We invoke the Learner to update the $\lambda$ every 10 steps. To speed up the training process, the policy parameters are evaluated for adding to AIM policy and export once per 100 steps. For the online learner, we use the OGD (\ref{ogd}) with learning rate as $\eta_t := 1/\sqrt{t}$ \cite{zinkevich2003online}.

\noindent{\textbf{Budget control.}} To make a fair comparison between conventional methods and the offline constrained RL methods, we control the immediate budget spent to the same $3$ yuan using a linear programming task. We solve the task by taking the Q values of the RL method as a response estimation of the users, then taking the two-step framework, and adding a constraint until the cost of the DNN methods and the RL methods are the same.

\noindent{\textbf{Offline evaluation.}} To evaluate different methods on ground truth, a straightforward way is to evaluate the learned policy through an online A/B test. However, it may hurt user experiences and be prohibitively expensive in an online marketing campaign. As suggested by \cite{zou2020pseudo}, we use a clipped importance sampling method to evaluate the policy. To ensure a small variance and control the bias, the weight is clipped to $20$.

\subsection{Offline Experimental Results}

% \textcolor{red}{7day average DAU is used as long-term metric, which may not be long enough to change user behavior.}

\noindent{\textbf{Overall performance comparison.}} Results by an average of 5 repetitive experiment runs are obtained, and we report the three metrics in Table \ref{table_compare}. For each method, we report its results in terms of the average coupon redemption rate (1d DAU), the average DAU over 3 days (3d DAU), and over 7 days (7d DAU). The redemption rate measures the immediate effect of the marketing campaign, and the average DAU over 3 days and 7 days measure the long-term effect of running the marketing campaign. From Table \ref{table_compare} we have following observations: \textbf{(1)} Though the conventional method yields a nice result in optimizing short-term metrics (1d DAU), they are outperformed by the offline constrained RL methods when considering the long-term effect (3/7d DAU). \textbf{(2)} Though lacks theoretical guarantee, using a single Q-network outperforms the AIM-mean method yield by the game-theoretic approach. Meanwhile, both are consistently outperformed by the AIM-greedy method, which verifies our theoretical analysis. Note that according to corollary \ref{thm4_2}, the measurement vector of the original mixed policy method is exactly the same as the AIM-mean method, and hence they have identical performance.

% \textbf{(3)} The usage of EDRR and BCQ improve the performance of the DDQN method significantly.

% \textcolor{red}{Figures and Algorithms appears without explanation. Fig. 1 and Algorithm 4 are not mentioned anywhere it the paper.}

% \setlength{\belowcaptionskip}{-10pt}

\begin{figure}[tb] 
\begin{center}
\includegraphics[width=0.45\textwidth]{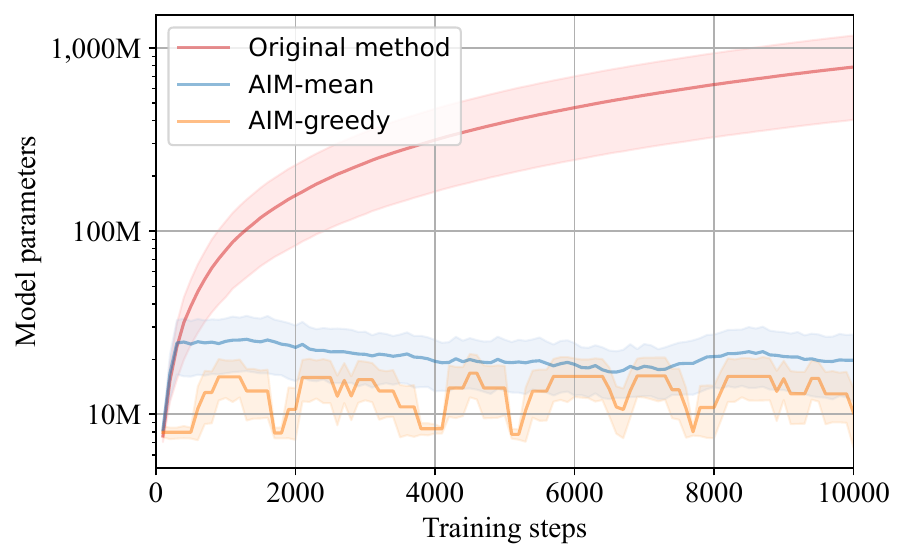}
\end{center}
\vspace{-4mm}
\caption{ The number of stored parameters in original mixed policy and AIM-based methods during the training process.
} \label{policy_efficiency}
\vspace{-2mm}
\end{figure}

\noindent{\textbf{The policy efficiency.}} The memory usage of the three offline constrained RL methods (using BCQ) with the standard deviation over the 5 repetitive runs is reported in Figure \ref{policy_efficiency} in terms of the number of parameters stored during the training process on full dataset. Directly applying the original mixed policy approach (Algorithm \ref{alg:algorithm_1}) suffers from an impractical memory cost. In contrast, our proposed methods significantly reduces memory cost to at most a constant, which are favorable for large-scale industrial scenarios.

\subsection{Online Experimental Results}

\begin{figure}[tb] 
\begin{center}
\includegraphics[width=0.5\textwidth]{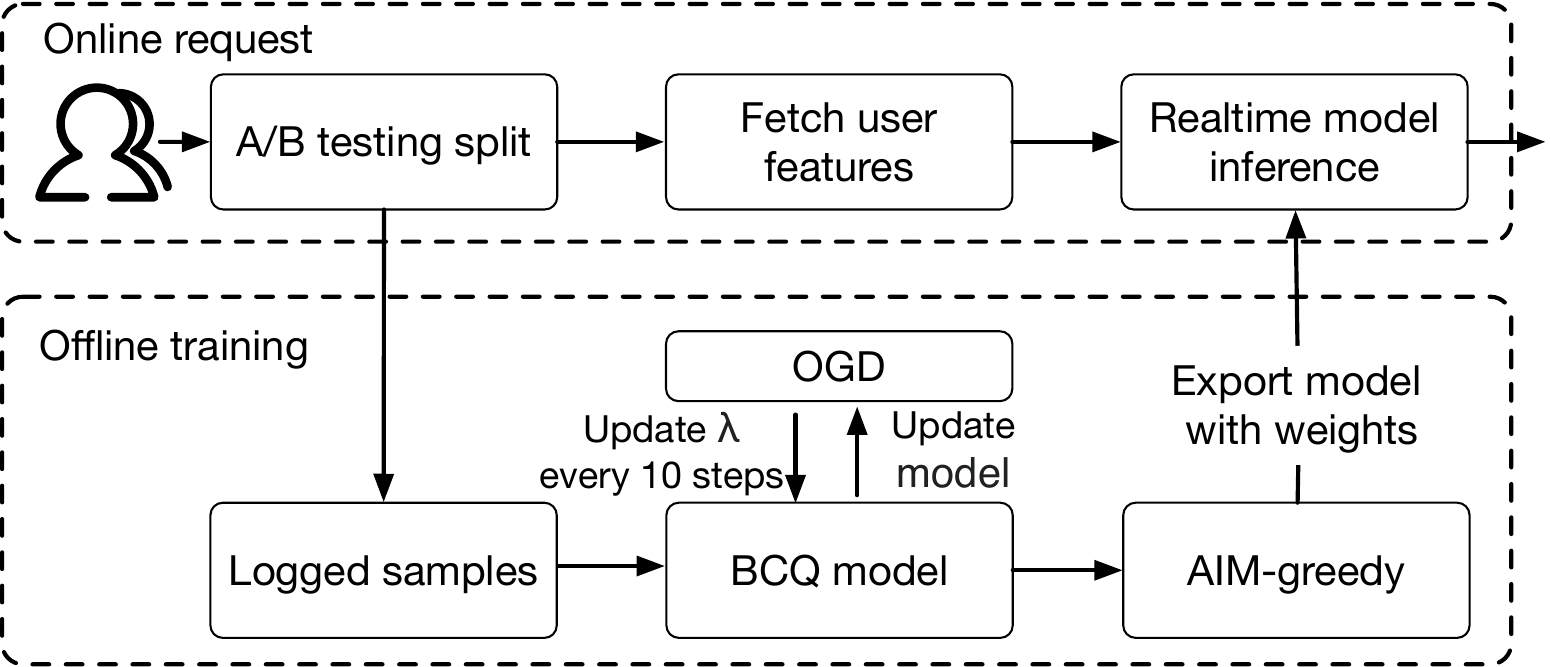}
\end{center}
\caption{The online request and offline training procedure. %\textcolor{red}{move to section 3 to let the readers have a holistic assessment of this approach.}
} 
\vspace{-2mm}
\label{deploy}
\end{figure}

We conduct a series of A/B experiments in a live system serving tens of millions of users and handing out more than one billion yuan worth cash coupons. Figure \ref{deploy} illustrates the online request serving process and the offline training procedure. We train the model using BCQ, where the $\lambda$ is updated by an OGD method. The model is evaluated by an AIM-greedy method, and if the current policy shall be added into the mixed policy, the model parameters will be exported to disk, together with the weight for all exported models. These models are then loaded to our online serving machines, to provide online budget allocation services.

The online experiment is conducted by performing user-level randomization. We start an online experiment with a small percentage of traffic and gradually increase the traffic. The offline constrained deep RL model is compared with the baseline model that uses DIN with a linear programming solver to find the allocation plan. The online experiment run for a week, with 5.5 million cash coupons sent by each of the models. Though the offline constrained RL method has a lower coupon redemption rate (1d DAU) than the previous DIN model, after running the experiment for a week, it improves the weekly average DAU by 0.6\% (7d DAU) for the app. Meanwhile, it reduces the coupon cost by 0.8\%. Our retrospective analysis shows that the emphasis on long-term value drives this model to allocate more budgets to inactive users, and accelerate the process of user acquisition. The more frequent retention of these inactive users compensates for the short-term metrics deterioration of active users. The model is then successfully deployed to the full traffic during the marketing campaign after a week's A/B testing.

% \begin{algorithm}[tb]
% \caption{Deploying Offline Constrained RL with AIM policy}
% \label{alg:instance}
% \begin{flushleft}
% \textbf{Input}: Maintain policy algorithm AIM-greedy(), pre-collected dataset $D := \{(s_i, a_i, s_i', r_i, \vc_i)\}_{i=1}^n$, offline RL algorithm $\mathtt{BCQ}(\pi, D)$ that updates policy $\pi$ using dataset $D$.
% % \textbf{Output}: AIM policy $u_t$
% \end{flushleft}
% %
% \begin{algorithmic}[1] %[1] enables line numbers
% \FOR{$t=1,2,\dots, T$}
% \STATE $\pi_t \gets \pi_{t-1}$ \COMMENT{warm-start }
% \FOR[$\mathtt{Best\_response}(\vlambda_{t})$ ]{$k=1,2,\dots, K$}
% \STATE Sample a batch $\{(s, a, s', r, \vc)\}$ from $D$
% \STATE Calculate reward $r':= r + \vlambda_t * (\vc - \tau)$
% \STATE $\pi_t \gets \mathtt{BCQ}(\pi_t, \{(s, a, s', r')\})$ \COMMENT{training}
% \ENDFOR
% \STATE Evaluate $J_\vc(\pi_t)$ using an extra Q-network
% \STATE $\vlambda_t \gets \mathtt{OGD}(\vlambda_{t-1}, J_\vc(\pi_t)) $
% \STATE $\mu \gets $ AIM-greedy$(\pi_t)$
% \ENDFOR
% \STATE \textbf{Return} $\mu$
% \end{algorithmic}
% \end{algorithm}

\section{Conclusions}

% We introduce an offline constrained RL approach to optimize marketing budget allocation. 

We show the difficulty of marketing budget allocation tasks using offline data from a real-world large-scale campaign, and propose using offline constrained deep RL techniques to deal with the long-term effect. Existing methods require randomizing among infinitely many policies, which suffers from an unacceptably high memory cost for large-scale scenarios. To tackle this issue, we propose AIM-mean and AIM-greedy algorithms, which reduce the memory consumption to a constant. In addition, we show that AIM-mean is guaranteed to convergence to the optimal policy, and AIM-greedy performs no worse than any single-best policy. The theoretical analyses are verified in a real-world marketing campaign, where the AIM-greedy is shown to outperform all other methods and is successfully deployed to all traffic in the campaign.

% \begin{table}
%   \caption{Frequency of Special Characters}
%   \label{tab:freq}
%   \begin{tabular}{ccl}
%     \toprule
%     Non-English or Math&Frequency&Comments\\
%     \midrule
%     \O & 1 in 1,000& For Swedish names\\
%     $\pi$ & 1 in 5& Common in math\\
%     \$ & 4 in 5 & Used in business\\
%     $\Psi^2_1$ & 1 in 40,000& Unexplained usage\\
%   \bottomrule
% \end{tabular}
% \end{table}

% To set a wider table, which takes up the whole width of the page's
% live area, use the environment \textbf{table*} to enclose the table's
% contents and the table caption.  As with a single-column table, this
% wide table will ``float'' to a location deemed more
% desirable. Immediately following this sentence is the point at which
% Table~\ref{tab:commands} is included in the input file; again, it is
% instructive to compare the placement of the table here with the table
% in the printed output of this document.

% \begin{table*}
%   \caption{Some Typical Commands}
%   \label{tab:commands}
%   \begin{tabular}{ccl}
%     \toprule
%     Command &A Number & Comments\\
%     \midrule
%     \texttt{{\char'134}author} & 100& Author \\
%     \texttt{{\char'134}table}& 300 & For tables\\
%     \texttt{{\char'134}table*}& 400& For wider tables\\
%     \bottomrule
%   \end{tabular}
% \end{table*}

% Always use midrule to separate table header rows from data rows, and
% use it only for this purpose. This enables assistive technologies to
% recognise table headers and support their users in navigating tables
% more easily.

%%
%% The next two lines define the bibliography style to be used, and
%% the bibliography file.
\bibliographystyle{ACM-Reference-Format}
\balance
\bibliography{samples/sample-base}

\end{document}